\documentclass[letterpaper]{article} 

\usepackage{aaai2027}

\nocopyright

\usepackage[hyphens]{url}
\usepackage{graphicx}
\usepackage{natbib}
\usepackage{caption}

\usepackage{algorithm}
\usepackage{algorithmic}

\usepackage{adjustbox}
\usepackage{colortbl}
\usepackage{booktabs}
\usepackage[table,xcdraw]{xcolor}
\usepackage{amsmath}
\usepackage{amssymb}
\usepackage{multirow}
\usepackage{makecell}

\usepackage{array}
\usepackage{xspace}

\usepackage{subcaption}

\usepackage[breakable,skins]{tcolorbox}
\definecolor{tcolorboxblue}{rgb}{0.2, 0.3, 0.6}
\definecolor{tcolorboxpink}{rgb}{1.0, 0.75, 0.8}
\definecolor{democraticblue}{RGB}{0,174,243}
\definecolor{republicanred}{RGB}{232,27,35}

\usepackage{listings}

\colorlet{jsonPunct}{red!60!black}
\definecolor{jsonBackground}{HTML}{EEEEEE}
\definecolor{jsonDelim}{RGB}{20,105,176}
\colorlet{jsonNumb}{magenta!60!black}
\lstdefinelanguage{json}{
    basicstyle=\normalfont\ttfamily,
    showstringspaces=false,
    breaklines=true,
    frame=lines,
    backgroundcolor=\color{jsonBackground!29},
    literate=
     *{0}{{{\color{jsonNumb}0}}}{1}
      {1}{{{\color{jsonNumb}1}}}{1}
      {2}{{{\color{jsonNumb}2}}}{1}
      {3}{{{\color{jsonNumb}3}}}{1}
      {4}{{{\color{jsonNumb}4}}}{1}
      {5}{{{\color{jsonNumb}5}}}{1}
      {6}{{{\color{jsonNumb}6}}}{1}
      {7}{{{\color{jsonNumb}7}}}{1}
      {8}{{{\color{jsonNumb}8}}}{1}
      {9}{{{\color{jsonNumb}9}}}{1}
      {`}{{{\color{jsonNumb}`}}}{1}
      {'}{{{\color{jsonNumb}'}}}{1}
      {"}{{{\color{jsonNumb}"}}}{1}
      {:}{{{\color{jsonPunct}{:}}}}{1}
      {,}{{{\color{jsonPunct}{,}}}}{1}
      {\{}{{{\color{jsonDelim}{\{}}}}{1}
      {\}}{{{\color{jsonDelim}{\}}}}}{1}
      {[}{{{\color{jsonDelim}{[}}}}{1}
      {]}{{{\color{jsonDelim}{]}}}}{1},
}

\usepackage{cleveref}

\title{
The Cost of Knowing:
A Resource-Aware Protocol for Benchmarking Hallucination
Beyond Static Leaderboards
}

\author{
    Keyu Li,\textsuperscript{\rm 1}
    Jin Gao,\textsuperscript{\rm 1}
    Dequan Wang\textsuperscript{\rm 1,*}
}

\affiliations{
    \textsuperscript{\rm 1}
    Shanghai Jiao Tong University\\
    Shanghai, China\\
    \{chlorophyll, gaojin, dequanwang\}@sjtu.edu.cn
}

\begin{document}

\maketitle

\begingroup
\renewcommand{\thefootnote}{*}
\footnotetext{Corresponding author.}
\endgroup

\begin{abstract}
On standard factuality tasks, frontier models now cluster near the top of the scale. The question is therefore shifting from how factual a system is toward how much compute that factuality costs. Static leaderboards score factuality in isolation and treat compute as free, so they cannot tell a genuinely better system apart from one that simply spends more. Consider a ranking reversal. A brute-force Best-of-4 agent posts the higher raw factuality score (H-Score 0.9169 vs.\ 0.9103) and would top a static leaderboard, but once cost is counted it is the worse system, losing on Q-Score (0.5169 vs.\ 0.5217) at roughly four times the tokens and latency, under a reported cost weight whose sensitivity we sweep. So the system that tops a static leaderboard can be the worse one to deploy. To make this trade-off visible, we introduce MAS-HQ (\textbf{M}ulti-\textbf{A}gent \textbf{S}ystem \textbf{H}allucination \textbf{Q}uest), a resource-aware evaluation protocol. It wraps any factuality detector and normalizes for cost, and it pits systems against each other rather than scoring them in isolation. The Q-Score measures factuality minus normalized cost under a competitive match. Across summarization and open-domain QA, single-agent baselines drift into resource-heavy over-optimization, while competition elicits more resource-efficient policies. These gains are small but consistent, and stable across 100 trials. The axis stays discriminative for frontier systems (Gemini-2.5-Pro, and GPT-5) whose raw factuality scores are already bunched near the ceiling. MAS-HQ provides a reproducible way to measure how much a factual answer costs.
\end{abstract}

\begin{figure*}[h]
    \centering
\includegraphics[width=0.9\linewidth]{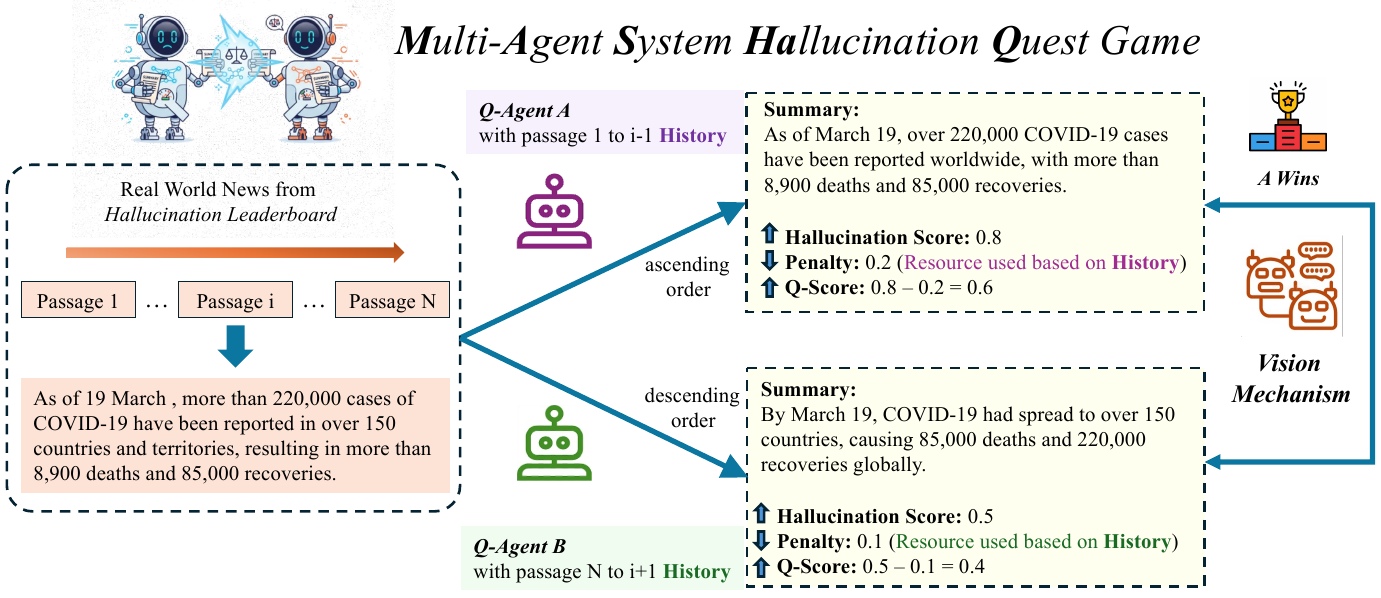}

\caption{
\textbf{Same tasks, reversed ranking: higher factuality, worse once cost is counted.} Two Q-Agents compete on MAS-HQ; the original passage list is input in different sequences to Q-Agent A and Q-Agent B, which devise strategies to summarize each passage using mechanisms such as the vision mechanism. Hallucination scores and resource-usage penalties are recorded and combined into the Q-Score, and the agent with the higher Q-Score wins---making the cost-of-knowing trade-off visible rather than hidden. In the current context, Q-Agent A wins.}
\label{fig:teaser}
\end{figure*}

\section{Introduction}

On standard factuality tasks, frontier LLMs now cluster near the top of the scale~\citep{huang2025survey, alzahrani2024benchmarks,huang2025exploring,singh2025leaderboard}. As they do, the informative question moves from \emph{how factual} a system is to \emph{how much compute that factuality costs}. Static leaderboards still score factuality in isolation and treat compute as free. Under that convention a model can ``buy'' a higher score through computationally extravagant mitigation, such as extensive Best-of-N sampling or exhaustive reasoning chains, without penalty. So a static leaderboard cannot tell a genuinely better system from one that merely spends more.

Consider a ranking reversal. A brute-force Best-of-4 agent posts the higher raw factuality score (H-Score 0.9169 vs.\ 0.9103) and would top a static leaderboard. Once cost is counted it is the worse system: it loses on Q-Score (0.5169 vs.\ 0.5217) at roughly four times the tokens and latency, scored under a reported cost weight whose sensitivity we sweep, so the static-score winner can be the wrong deployment pick. The trade-off is not a small-model artifact, since it persists for frontier systems (Gemini-2.5-Pro, and GPT-5) whose raw factuality already sits near the ceiling. It is compounded in Multi-Agent Systems (MAS), where interactions can amplify and propagate hallucinations, outputs that appear plausible yet are factually erroneous~\citep{macpherson2013hallucination, huang2021factual,ji2023survey,li2022faithfulness,schmidgall2025agentrxiv}. In such settings latency and compute are first-order. In high-frequency trading or real-time news intelligence, an agent that delivers equally credible insights on half the budget delivers greater operational value.

To make this trade-off visible, we introduce MAS-HQ (\textbf{M}ulti-\textbf{A}gent \textbf{S}ystem \textbf{H}allucination \textbf{Q}uest), a resource-aware evaluation protocol built on standard information-synthesis tasks~\citep{leaderboard,hhem-2.1-open} as objective, quantifiable proxies. The accuracy term is \emph{metric-agnostic}: any factuality detector works, and its reliability bounds the scores. The Q-Score then subtracts normalized cost under a competitive match, so an accuracy gain bought with disproportionate compute is scored down. Each measurement pairs with fixed budgets and a competitive ranking rule, which surfaces the ranking changes static leaderboards hide. We instantiate the protocol with the Q-Agent, whose optional parameterized telemetry knob surfaces partial state about an opponent only when a costly action such as self-review is taken, mirroring the partial observability of latency, retries, and tool calls in deployment. \Cref{fig:teaser} gives an overview.

Experiments show that competition produces small but consistent shifts. Single-agent baselines fall into the \emph{over-optimization trap}, spending disproportionate resources for marginal accuracy; under a shared budget, agents adopt more resource-efficient policies. The gaps stay stable across 100 trials, and they change relative rankings that a static score cannot surface.

In summary, our contributions are: (1) \textbf{The MAS-HQ protocol:} a resource-aware evaluation protocol that wraps any factuality metric and scores accuracy against normalized operational cost, making the cost-of-knowing trade-off visible. (2) \textbf{A modular architecture for controlled ablations:} including an optional parameterized telemetry knob (the ``vision mechanism''), which isolates when competition reverses rankings versus when it collapses onto the static axis. (3) \textbf{Reproducible analysis:} experiments showing shared budgets and competition elicit resource-efficient behaviors, consistent across trials, and a ranking reversal that static leaderboards overlook.

\section{Related Work}

\paragraph{LLM Hallucination and Resource-Agnostic Evaluation}
Large Language Models~\citep{li2026agencybench,wu2025innovatorbench,xiao2025limi,li2025datasetresearch,li2026aligned,jiang2026davinci,gao2024kan} hallucination—the generation of fluent yet factually erroneous content arising from flawed internals, biased data, or misaligned external knowledge~\citep{macpherson2013hallucination, huang2025survey, huang2021factual, ji2023survey, li2022faithfulness, bender2021dangers, li2023batgpt, holtzman2019curious}—remains a critical challenge. Various mitigation strategies, ranging from retrieval-augmentation to complex reasoning chains, have subsequently emerged~\citep{abbas2021semdedup, dai2023neural, gao2022rarr, shi2023context, chuang2023dola}. However, existing benchmarks~\citep{lin2021truthfulqa, li2024dawn, leaderboard, hhem-2.1-open, cheng2023evaluating} predominantly focus on static, resource-agnostic hallucination reduction, inadvertently rewarding computationally extravagant approaches (e.g., exhaustive Best-of-N sampling) while ignoring operational costs like latency and API calls. MAS-HQ addresses this gap by explicitly evaluating efficiency alongside veracity, penalizing the brute-force accumulation of factuality.

\begin{figure*}[t]
    \centering
\includegraphics[width=0.9\linewidth]{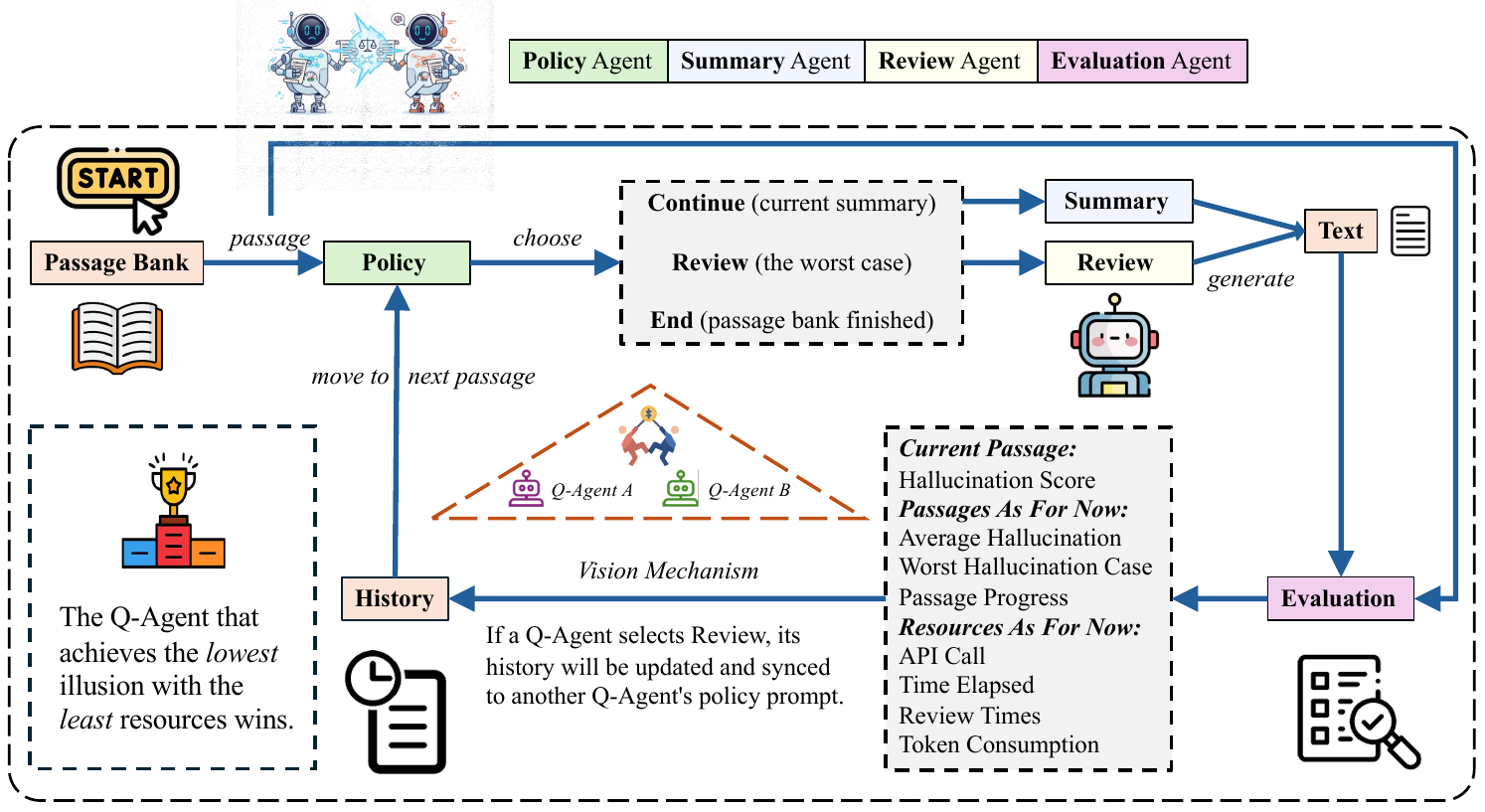}

\caption{
\textbf{Overview of the composition of the Q-Agent.} Policy Agent receives both the Q-Agent's own historical information and the opponent's historical information (via the vision mechanism) to determine the next strategy. The text summary generated by either the Summary Agent or the Review Agent is evaluated by the Evaluation Agent, which computes the hallucination score (H-Score) and resource usage. After all steps are completed, the overall Q-Score is calculated, which includes the average hallucination score and resource consumption penalties.
}
\label{fig:method}
\end{figure*}

\paragraph{Multi-Agent Systems and Economic Constraints}
As Multi-Agent Systems (MAS) deploy autonomous agents capable of complex interaction across increasingly diverse domains~\citep{liu2025advances, guo2024large, rasheed2024can, wang2024xuat, huang2023agentcoder, hong2023metagpt, chen2023multi, yu2023co, ke2024enhancing, ni2024mechagents, xie2024can, jiang2024large, wu2023deciphering, fan2024can, zhang2024generative, zhang2024agentcf}, evaluating their decision quality under stringent operational budgets in time-sensitive environments becomes paramount. While current benchmarks~\citep{zhu2025multiagentbench} valuably assess open-ended collaboration, they often neglect the rigorous quantification of information fidelity against strict resource limitations. MAS-HQ addresses this by introducing a game-theoretic paradigm that utilizes standard generation tasks as quantifiable proxies, forcing agents to strategically balance task success against resource consumption and adversarial pressure.

\section{MAS-HQ Framework and Reference Architecture}

Static factuality leaderboards presuppose effectively infinite compute. They score how factual a system is while implicitly assuming the compute that buys that factuality is free, so they cannot separate a genuinely better system from one that merely spends more. We introduce MAS-HQ (\textbf{M}ulti-\textbf{A}gent \textbf{S}ystem \textbf{H}allucination \textbf{Q}uest), a controlled resource-aware protocol that makes rankings cost-honest and makes such reversals visible. MAS-HQ treats standard summarization and QA tasks as quantifiable proxies for information synthesis and scores factuality jointly against normalized operational cost, so speed, cost, and accuracy are all weighed rather than assumed free. Any factuality detector can serve as the accuracy term, so the protocol stays metric-agnostic. \Cref{fig:method} shows the MAS-HQ evaluation and the Q-Agent reference implementation.

\subsection{MAS-HQ Mechanics and the Q-Score Objective}
The Q-Score just adds a cost term to factuality: factuality minus normalized cost under a competitive match, not a replacement for the factuality metric. Built on established datasets~\citep{leaderboard,hhem-2.1-open}, MAS-HQ pairs agents that sequentially process information streams while trading factual consistency against operational cost, aggregated into a match-specific utility, the $Q\text{-}Score$:
$$ Q\text{-}Score = \frac{1}{N} \sum_{i=1}^{N} \left( \alpha \cdot H\text{-}Score_{i} - \beta \cdot P_i \right) $$
For each of $N$ tasks, $H\text{-}Score_{i} \in [0,1]$ is the factuality metric from the plugged-in detector, and $P_i$ is a normalized operational-cost penalty computed at evaluation time. This evaluation-time cost differs from the policy-side review cost an agent incurs internally when it refines. The Q-Score subtracts only the former, so cost-awareness is scored uniformly across agents and is not baked into the accuracy metric. The penalty $P_i$ aggregates token count (input plus output), API calls, review iterations, and wall-clock runtime, each max-normalized over the agents in the same match, which makes $P_i$ a relative rather than absolute cost. The parameters $\alpha, \beta > 0$ are deployment-SLA knobs encoding a particular accuracy-versus-cost operating point, so we report Q-Score sensitivity across a range of $\beta$. Because the penalty is relative to the co-present competitor, the $Q\text{-}Score$ is a match-specific payoff and not a standalone model score. Recovering an intrinsic per-agent ranking would require tournament or Elo-style aggregation over many matches, which we leave to future work.

\begin{table*}[t]

\centering
\renewcommand{\arraystretch}{1.2}
\adjustbox{max width=0.9\linewidth}{
\large
    \begin{tabular}{cccccccc}

    \toprule
        \multirow{2}{*}{\shortstack{\textbf{Q-Agent} \\ \textbf{Competition}}}
   & \multicolumn{6}{c}{\textbf{Metrics}}   \\
    \cmidrule(r){2-8} 
     & H-Score~$\uparrow$ & API Call~$\downarrow$ & Tokens~$\downarrow$ & Review~$\downarrow$ & Time~$\downarrow$&Q-Score~$\uparrow$&Winner  \\
\bottomrule

    A: GPT-4o-mini  &  \cellcolor{gray!20}0.9103 & \cellcolor{gray!20}\textbf{2417} & \cellcolor{gray!20}\textbf{1.36M} & \cellcolor{gray!20}\textbf{791}& \cellcolor{gray!20}\textbf{8.83k}& \cellcolor{gray!20}\textbf{0.5217}&\checkmark\\

    B: GPT-4o-mini  &  \textbf{0.9132} & 2438 & 1.44M &812 & 8.98k&0.5132&\\

     \bottomrule

    A: Qwen-Max  & \cellcolor{gray!20}\textbf{0.9030} & \cellcolor{gray!20}2304 & \cellcolor{gray!20}\textbf{1.48M} & \cellcolor{gray!20}682& \cellcolor{gray!20}\textbf{13.77k} &\cellcolor{gray!20}\textbf{0.5101}&\checkmark\\

    B: Qwen-Max  &0.8994 & \textbf{2264} & 1.52M &\textbf{642} & 14.45k&0.5070&\\

   \bottomrule

    A: Deepseek-V3  &  0.8860 & 2292 & 1.42M & 666& 12.26k&0.4860& \\

    B: Deepseek-V3  &  \cellcolor{gray!20}\textbf{0.8894} & \cellcolor{gray!20}\textbf{2233} & \cellcolor{gray!20}\textbf{1.38M} & \cellcolor{gray!20}\textbf{607}& \cellcolor{gray!20}\textbf{12.06k}&\cellcolor{gray!20}\textbf{0.5051}&\checkmark\\

               \bottomrule

    A: Gemini-2.0-Flash  &  \textbf{0.9026} & 2262 & 1.56M & 642&20.29k & 0.5026&\\

    B: Gemini-2.0-Flash  &  \cellcolor{gray!20}0.9016 &\cellcolor{gray!20} \textbf{2157} & \cellcolor{gray!20}\textbf{1.53M} & \cellcolor{gray!20}\textbf{537}& \cellcolor{gray!20}\textbf{19.82k}&\cellcolor{gray!20}\textbf{0.5273}&\checkmark\\

                    \bottomrule

    A: Grok-3-beta  &\textbf{0.9070} & 2376 & 1.37M & 750&10.71k & 0.5070&\\

    B: Grok-3-beta  & \cellcolor{gray!20}0.9049 &\cellcolor{gray!20} \textbf{2253} & \cellcolor{gray!20}\textbf{1.34M} & \cellcolor{gray!20}\textbf{627}&\cellcolor{gray!20}\textbf{10.15k} &\cellcolor{gray!20}\textbf{0.5337}&\checkmark\\

    \bottomrule
    \end{tabular}
}

\caption{
\textbf{The reversal recurs across LLMs: the highest-factuality agent is often not the winner.} Within each competition group, Q-Agents share the same underlying LLM. Once cost is scored, the winner (Q-Score) is repeatedly not the agent with the higher raw factuality (H-Score)---resource efficiency (API Calls, Tokens, Review, Time) decides the cost-honest ranking.
}

\label{tab:main_results_1}
\end{table*}

\begin{table*}[h]
\centering
\renewcommand{\arraystretch}{1.2}
\adjustbox{max width=0.95\linewidth}{
\large
    \begin{tabular}{lcccccc}
    \toprule
        \textbf{Setup (Model: GPT-4o-mini)} & H-Score~$\uparrow$ & API Call~$\downarrow$ & Tokens~$\downarrow$ & Review~$\downarrow$ & Time (s)~$\downarrow$ & Q-Score~$\uparrow$ \\
    \midrule
        Single Agent: Best-of-1 & 0.9144 & 2477 & 1.62M & 851 & 9.12k & 0.5144 \\
        Single Agent: Best-of-2 & \cellcolor{gray!20}0.9167 & \cellcolor{gray!20}5201 & \cellcolor{gray!20}3.40M & \cellcolor{gray!20}1787 & \cellcolor{gray!20}19.15k & \cellcolor{gray!20}0.5167 \\
        Single Agent: Best-of-4 & \textbf{0.9169} & 9660 & 6.32M & 3310 & 35.57k & 0.5169 \\
    \midrule
        \textbf{Multi-Agent: Q-Agent A (Ours)} & 0.9103 & \textbf{2417} & \textbf{1.36M} & \textbf{791} & \textbf{8.83k} & \textbf{0.5217} \\
        Multi-Agent: Q-Agent B (Ours) & 0.9132 & 2438 & 1.44M & 812 & 8.98k & 0.5132 \\
    \bottomrule
    \end{tabular}
}
\caption{
\textbf{The ranking reversal.} Best-of-4 wins on raw factuality (H-Score 0.9169) but loses on Q-Score (0.5169 vs. 0.5217) at $\sim$4$\times$ the cost: the static-score winner is here the worse deployment pick. This is the over-optimization trap---MAS-HQ agents adopt more resource-efficient empirical policies, capturing most of the factuality at a fraction of the cost.
}
\label{tab:best_of_n}
\end{table*}

\paragraph{The Q-Agent Reference Architecture.} As a transparent, reproducible baseline, we provide the Q-Agent, a modular reference implementation (analogous to ReAct~\citep{yao2022react}) of four components. The \textbf{Policy Agent (PA)} decides the next action (\texttt{continue}, \texttt{review}, or \texttt{end}) from the current state and any available competitor telemetry. The \textbf{Summarization Agent (SA)} executes the primary task. To avoid the degenerate ``say less to avoid being wrong'' shortcut, we deliberately isolate it from the Q-Score, which confines cost-awareness to the PA. The \textbf{Review Agent (RA)} refines summaries at an additional review cost, while the \textbf{Evaluation Agent (EA)} scores quality~\citep{hhem-2.1-open} and tracks consumption to inform the PA.

\subsection{Competitive Dynamics and Optional Telemetry}
We instantiate MAS-HQ as a head-to-head match to remove the infinite-compute assumption, with Agent A and Agent B processing tasks in opposite directions to prevent trivial convergence.

\paragraph{The Vision Mechanism.} MAS-HQ includes an optional, parameterized telemetry knob controlling how much of a competitor's activity is observable, mirroring the partial telemetry (latency, retries, tool-calls) available in real deployments. When it is enabled and an agent executes a costly action such as \texttt{review}, a snapshot of its state ($V_{\text{A},i}^{\text{opponent}}$) is disclosed to its rival, so the policy is conditioned on observed competitor cost:
$$
\text{Choice}_{\text{B},i} = \text{Policy}_{\text{B}}(T_{\text{B},i}, V_{\text{B},i}^{\text{self}}, V_{\text{A},i}^{\text{opponent}}), \quad \text{for} \quad i = 1, \dots, N
$$
Turning this knob lets us measure how opponent-conditioned review behavior changes the accuracy--cost trade-off; we examine the full-versus-collapsed match ablation in appendix. A Limited Review Cycle budget ($R$) further constrains how much refinement an agent may spend. The information structure of a match (private state, conditional observation, sequential actions, and terminal utility) can be described compactly as a POSG. We use this only as descriptive vocabulary and do not solve for equilibria, which remains future work.

\begin{table*}[h]
\centering
\renewcommand{\arraystretch}{1.2}
\adjustbox{max width=0.9\linewidth}{
\large
    \begin{tabular}{lcccc}
    \toprule
        \textbf{SOTA Competition ($\alpha=1, \beta=0.01$)} & H-Score~$\uparrow$ & Review~$\downarrow$ & Resource Penalty~$\downarrow$ & Q-Score~$\uparrow$ \\
    \midrule
        Q-Agent A: GPT-5 & 0.9312 & 867 & 0.4000 & 0.5312 \\
        Q-Agent B: GPT-5 & \cellcolor{gray!20}\textbf{0.9278} & \cellcolor{gray!20}\textbf{834} & \cellcolor{gray!20}\textbf{0.3897} & \cellcolor{gray!20}\textbf{0.5381} (\checkmark) \\
    \midrule
        Q-Agent A: Gemini-2.5-Pro & \textbf{0.9264} & 859 & 0.4000 & 0.5264 \\
        Q-Agent B: Gemini-2.5-Pro & \cellcolor{gray!20}0.9221 & \cellcolor{gray!20}\textbf{824} & \cellcolor{gray!20}\textbf{0.3911} & \cellcolor{gray!20}\textbf{0.5310} (\checkmark) \\
    \midrule
        Q-Agent A: GPT-5 & \cellcolor{gray!20}\textbf{0.9315} & \cellcolor{gray!20}869 & \cellcolor{gray!20}0.4000 & \cellcolor{gray!20}\textbf{0.5315} (\checkmark) \\
        Q-Agent B: GPT-4o-mini & 0.9078 & \textbf{846} & \textbf{0.3877} & 0.5201 \\
    \bottomrule
    \end{tabular}
}
\caption{
\textbf{The cost-of-knowing trade-off persists where raw factuality saturates.} For frontier systems (GPT-5; Gemini-2.5-Pro) whose H-Scores already cluster near the top of the scale, the axis stays discriminative: higher pure factuality (H-Score) does not guarantee a win once the resource penalty is scored, so the reversal is not a small-model artifact.
}
\label{tab:sota_results}
\end{table*}
\begin{table*}[ht]
\centering
\renewcommand{\arraystretch}{1.2}
\adjustbox{max width=0.95\linewidth}{
\large
    \begin{tabular}{ccccccc}
    \toprule
        $\alpha$ & $\beta$ & \textbf{Model Setup} & H-Score~$\uparrow$ & Review Count & Resource Penalty & Q-Score \\
    \midrule
        1 & 0 & Q-Agent A: GPT-4o-mini & 0.9166 & 854 & 0.3899 & 0.5267 \\
        1 & 0 & Q-Agent B: GPT-4o-mini & \textbf{0.9189} & 889 & 0.4000 & \textbf{0.5189} \\
    \midrule
        1 & \textbf{0.01} & \textbf{Q-Agent A: GPT-4o-mini (Main)} & 0.9103 & \textbf{791} & \textbf{0.3886} & \textbf{0.5217} \\
        1 & \textbf{0.01} & \textbf{Q-Agent B: GPT-4o-mini (Main)} & \textbf{0.9132} & 812 & 0.4000 & 0.5132 \\
    \midrule
        1 & 0.02 & Q-Agent A: GPT-4o-mini & 0.9043 & \textbf{755} & 0.3901 & \textbf{0.5142} \\
        1 & 0.02 & Q-Agent B: GPT-4o-mini & \textbf{0.9096} & 772 & \textbf{0.4000} & 0.5096 \\
    \bottomrule
    \end{tabular}
}
\caption{
\textbf{$\beta$ toggles between static and cost-honest ranking.} Fixing $\alpha=1$ and sweeping $\beta$. At $\beta=0$ the protocol collapses onto the saturating raw-accuracy axis (cost unscored, infinite resources), while $\beta=0.01$ is an intermediate deployment-SLA setting that balances quality against cost; we report sensitivity rather than claim a canonical value.
}
\label{tab:hyper_ablation}
\end{table*}

\section{Experiments}

Our experiments ask what a static factuality score cannot resolve. First, does jointly scoring factuality against normalized operational cost expose accuracy--cost trade-offs, up to a reversal of the ranking, that static infinite-compute leaderboards hide? Second, do agents, including frontier models, adopt measurably different resource-efficient behaviors when cost is scored? And once raw factuality already clusters near the top of the scale, does the axis stay discriminative and reproducible, and does it depend on the choice of metric?

\subsection{Experimental Setup and Metric Validation}
\label{subsec:setup}

Evaluations run within the MAS-HQ protocol on over 1,000 long-form news passages. Factual consistency ($H\text{-}Score$) is computed by a pre-trained discriminator~\citep{hhem-2.1-open} from the hallucination leaderboard~\citep{leaderboard}; the protocol is metric-agnostic, and any detector can be substituted. To verify this automated metric, we run a blind human evaluation on 100 randomly sampled summaries prior to large-scale testing. It yields a Pearson correlation of 0.96 between expert judgments and the discriminator's $H\text{-}Score$, which confirms its reliability as a ranking mechanism.

For a fair head-to-head comparison, we exclude passages that trigger safety refusals in any evaluated LLM, scoring competing agents only on the exact intersection both can process (808--813 passages per match). We set the hallucination threshold at $T = 0.85$, maximum review cycles at $R = 3$, and the $Q\text{-}Score$ cost weights at $\alpha = 1, \beta = 0.01$. Per the protocol, the penalty term $P_i$ max-normalizes API calls, total token consumption (input plus output), review iterations, and wall-clock runtime over the agents in each match. Here $\alpha$ and $\beta$ are deployment-SLA parameters, and we report sensitivity to them below. Q-Agent A processes the passage queue sequentially, Q-Agent B in reverse.

\subsection{A Ranking Reversal Under Best-of-N}
\label{subsec:best_of_n}

A static leaderboard scores factuality in isolation and assumes compute is free, so it rewards brute-force scaling. Models can ``buy'' accuracy by spending more, and the leaderboard cannot separate a genuinely better system from one that merely spends more. The clearest case is a ranking reversal, where the system that would top a static leaderboard is the one a deployer should not pick. We benchmark Q-Agents against a standard Single-Agent \textbf{Best-of-N} baseline (\Cref{tab:best_of_n}).

The reversal is direct. The resource-agnostic Best-of-4 strategy posts the higher $H\text{-}Score$ (0.9169 vs. 0.9103 for Q-Agent A) and would win a static leaderboard, but it buys those fractions of a point at nearly four times the token volume and latency. Once operational cost is scored, that spend does not pay off: Best-of-4 yields the \emph{lower} final $Q\text{-}Score$ (0.5169 vs. 0.5217). The static-score winner is the worse deployment pick, the over-optimization trap the protocol targets, with the Q-Agent capturing most of the factuality at a fraction of the cost. To check that this modest gap is not noise, we repeat the Q-Agent A vs. Q-Agent B match over 100 independent trials. Q-Agent A's advantage is stable (0.5216$\pm$0.0009 vs. 0.5131$\pm$0.0009), so the small $Q\text{-}Score$ gaps reflect a reproducible trade-off rather than run-to-run variance.

\begin{figure*}[t]
    \centering
\includegraphics[width=0.9\linewidth]{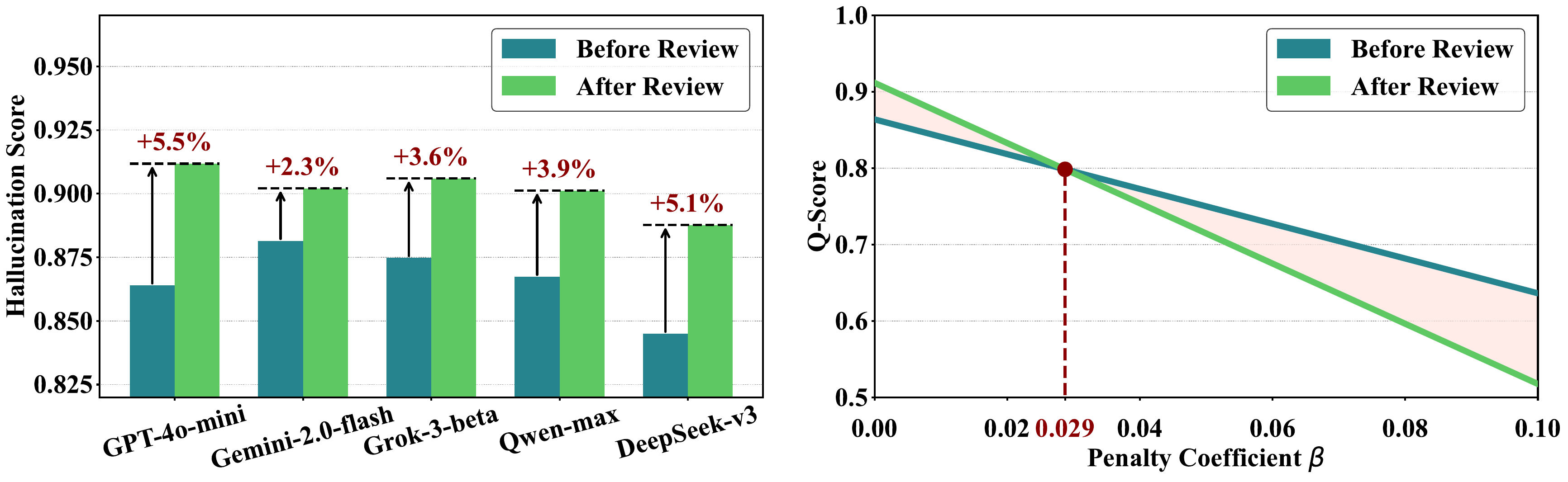}

\caption{
\textbf{Ablation study of the Review Agent in Q-Agent.} Each experiment group presents the average results over Q-Agent A and Q-Agent B. \textit{Left:} The Review Agent improves the H-Score and reduces hallucination to varying extents across different Q-Agent models. \textit{Right:} When both Q-Agents are built using GPT-4o-mini, the smaller the penalty coefficient $\beta$, the more the review pays off in the final Q-Score---as $\beta\to0$ the ranking slides back toward the static raw-accuracy axis.
}
\label{fig:review}
\end{figure*}

\subsection{Resource-Efficient Behavior and Frontier-Model Comparisons}
\label{subsec:competition_results}

We run head-to-head matches across a diverse spectrum of LLMs to examine how scored cost reshapes rankings. The reversal is not confined to Best-of-N. Across brackets in \Cref{tab:main_results_1}, the match winner is often the agent with the lower $H\text{-}Score$; in the GPT-4o-mini bracket, the agent with the lower factuality score wins on the final $Q\text{-}Score$ through greater frugality. \Cref{tab:main_results_2} further investigates cross-model dynamics and processing order, showing how agents on different foundation models (e.g., GPT-4o-mini vs. Grok-3-beta) leverage distinct efficiency profiles to change the outcome.

The harder test is whether the axis stays informative for the strongest systems. We scale evaluations to state-of-the-art (SOTA) systems, including \textbf{GPT-5} and \textbf{Gemini-2.5-Pro} (\Cref{tab:sota_results}), whose baseline $H\text{-}Scores$ already cluster near the top of the scale. The cost-of-knowing trade-off persists for these systems too: even they must decide whether a marginal factual refinement justifies its added resource penalty, and once raw scores cluster near the top, the cost-honest ranking is what still separates them. The effect holds for larger frontier systems, not just small models.

\subsection{Protocol Robustness, Scalability, and Ablations}

We conduct ablations isolating the protocol's core components.

\paragraph{Cost-Weight Sensitivity ($\alpha, \beta$).} Since $\alpha$ (quality) and $\beta$ (cost) are deployment-SLA parameters, we report a sensitivity analysis instead of fixing a single setting. \Cref{tab:hyper_ablation} details the sweep with $\alpha=1$ across varying $\beta$. Here $\beta$ toggles between static and cost-honest ranking. At $\beta=0$, with cost unscored, the protocol collapses onto the raw-accuracy axis, review counts surge, and the reversal disappears, reverting to the infinite-compute setting. At $\beta=0.02$ review cost dominates and suppresses $H\text{-}Scores$. The intermediate $\beta=0.01$ balances the two and reflects a realistic operational constraint; practitioners should pick $\alpha,\beta$ to match their own latency and token budgets. \Cref{fig:review} visualizes how adjusting $\beta$ shifts the final $Q\text{-}Score$ optimization path.

\begin{table}[h]

\centering
\renewcommand{\arraystretch}{1.2}
\adjustbox{max width=\linewidth}{
\begin{tabular}{lccccccc}
\toprule
\multirow{2}{*}{\shortstack{\textbf{Q-Agent} \\ \textbf{Competition}}} & \multicolumn{6}{c}{\textbf{Metrics}} \\
\cmidrule(r){2-8}
& H-Score $\uparrow$ & API Calls $\downarrow$ & Tokens $\downarrow$ & Reviews $\downarrow$ & Time (s) $\downarrow$ & Q-Score $\uparrow$ & Winner \\
\midrule
A: GPT-4o-mini & \cellcolor{gray!20}0.9108 & \cellcolor{gray!20}\textbf{2424} & \cellcolor{gray!20}\textbf{1.36M} & \cellcolor{gray!20}\textbf{798} & \cellcolor{gray!20}\textbf{8.84k} & \cellcolor{gray!20}\textbf{0.5214} & \cellcolor{gray!20}\checkmark \\
B: GPT-4o-mini & \textbf{0.9139} & 2445 & 1.44M & 819 & 8.99k & 0.5139 & \\
C: GPT-4o-mini & 0.9123 & 2434 & 1.40M & 808 & 8.89k & 0.5180 & \\
\bottomrule
\end{tabular}
}

\caption{
    \textbf{Scalability to 3-Agent Competition.} Results using GPT-4o-mini demonstrate that the MAS-HQ framework scales to N-player scenarios, where multi-lateral cost pressure remains measurable. Q-Agent A wins via resource efficiency despite having the lowest H-Score.
}
\label{tab:3_agent_results}

\end{table}
\begin{table}[h]

\centering
\renewcommand{\arraystretch}{1.2}
\adjustbox{max width=0.9\linewidth}{
\large
    \begin{tabular}{cccccccc}

    \toprule
        \multirow{2}{*}{\shortstack{\textbf{Q-Agent} \\ \textbf{Competition}}}
   & \multicolumn{6}{c}{\textbf{Metrics}}   \\
    \cmidrule(r){2-8} 
     & H-Score~$\uparrow$ & API Call~$\downarrow$ & Tokens~$\downarrow$ & Review~$\downarrow$ & Time~$\downarrow$&Q-Score~$\uparrow$&Winner  \\
\bottomrule

    A: GPT-4o-mini  & \textbf{0.9105} & 2401 & 1.34M &785 &\textbf{6.54k} & 0.5401&\\

    B: Grok-3-beta  & \cellcolor{gray!20}0.9035 & \cellcolor{gray!20}\textbf{2177} &\cellcolor{gray!20} \textbf{1.30M} &\cellcolor{gray!20}\textbf{561} & \cellcolor{gray!20}9.29k&\cellcolor{gray!20}\textbf{0.5445}&\checkmark\\

                    \bottomrule

    A: Grok-3-beta  & 0.9036 & \textbf{2312} & \textbf{1.34M} & \textbf{696}&10.08k &0.5278& \\

    B: GPT-4o-mini  & \cellcolor{gray!20}\textbf{0.9092} & \cellcolor{gray!20}2422 & \cellcolor{gray!20}1.43M &\cellcolor{gray!20}806 &\cellcolor{gray!20}\textbf{6.78k} &\cellcolor{gray!20}\textbf{0.5419}&\checkmark\\

    \bottomrule
    \end{tabular}
}

\caption{
\textbf{Main results investigating the impact of LLM and passage processing order.} In each competition group, Q-Agent A and Q-Agent B are constructed using different LLMs. Q-Agent A retrieves original text from the passage bank in forward order, while Q-Agent B retrieves in reverse order. The winning agent in each competition is highlighted with a light gray background, and better values for individual metrics are shown in \textbf{bold}.
}
\label{tab:main_results_2}
\end{table}

\paragraph{Architecture Agnosticism.} To rule out custom-engineering bias, we replace the Q-Agent with a standardized \textbf{ReAct Agent}~\citep{yao2022react} under identical rules (\Cref{tab:react_ablation}). The ReAct Agents produce nearly identical performance metrics, review frequencies, and $Q\text{-}Scores$. This indicates the resource-efficient behavior arises from the LLM's response to the scored-cost incentive rather than structural hardcoding, and that the protocol is agnostic to the agent scaffold.

\begin{figure*}[t]
    \centering
\includegraphics[width=0.9\linewidth]{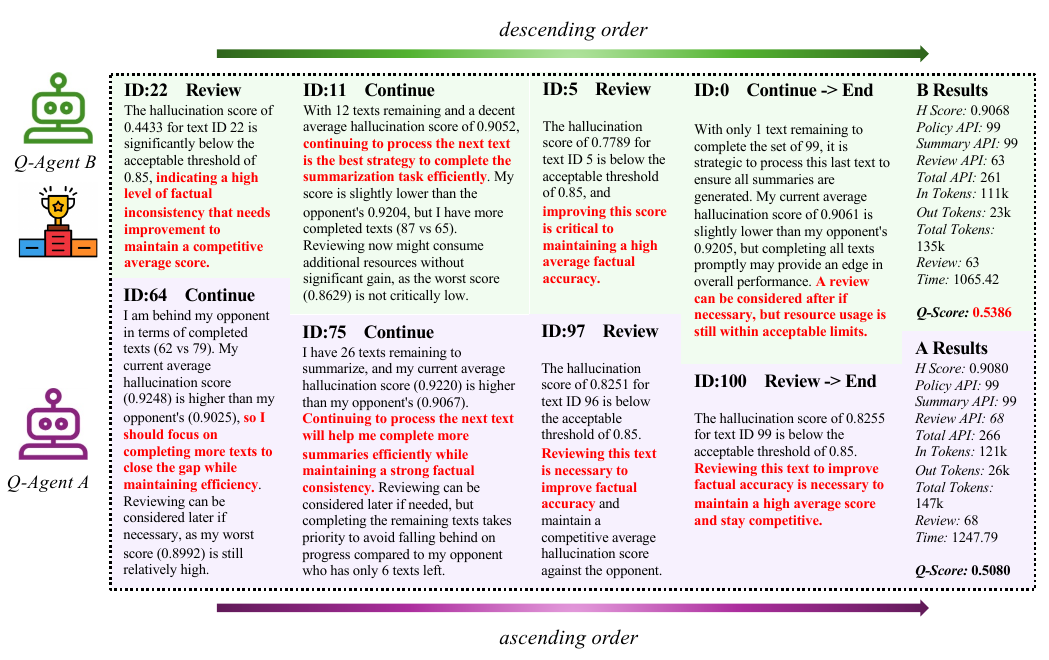}

\caption{
\textbf{Case study on a subset of MAS-HQ.} Through vision mechanism, Q-Agent A and Q-Agent B dynamically adjust their strategies based on their own H-Score, resource usage, and comparisons with the opponent, deciding whether to continue generating summaries or perform reviews. In this example the same reversal appears at the match level: Q-Agent B wins on the cost-honest Q-Score through lower resource usage despite a slightly lower H-Score. These are empirical, match-specific observations, not equilibrium claims.
}
\label{fig:case}
\end{figure*}

\begin{table}[h]
\centering
\renewcommand{\arraystretch}{1.2}
\adjustbox{max width=0.9\linewidth}{
\large
    \begin{tabular}{lcccc}
    \toprule
        \textbf{Architecture (GPT-4o-mini)} & H-Score~$\uparrow$ & Review Count & Resource Penalty & Q-Score~$\uparrow$ \\
    \midrule
        Q-Agent (Player A) & 0.9166 & 854 & 0.3899 & 0.5267 \\
        Q-Agent (Player B) & \textbf{0.9189} & 889 & 0.4000 & 0.5189 \\
    \midrule
        ReAct Agent (Player A) & 0.9166 & 855 & \textbf{0.3889} & \textbf{0.5277} \\
        ReAct Agent (Player B) & \textbf{0.9189} & \textbf{890} & 0.4000 & 0.5189 \\
    \bottomrule
    \end{tabular}
}
\caption{
\textbf{Architecture Agnostic Validation.} Replacing the custom Q-Agent with a standard ReAct Agent under MAS-HQ rules yields nearly identical behavioral metrics. This indicates that the resource-efficient behavior arises from the LLM's response to the scored-cost incentive, not from structural hardcoding.
}

\label{tab:react_ablation}

\end{table}
\begin{table}[h]

\centering
\renewcommand{\arraystretch}{1.2}
\adjustbox{max width=\linewidth}{
\begin{tabular}{cccccccc}
\toprule
\multirow{2}{*}{\shortstack{\textbf{Q-Agent} \\ \textbf{Competition}}} & \multicolumn{6}{c}{\textbf{Metrics}} \\
\cmidrule(r){2-8}
& \shortstack{H-Score $\uparrow$} & API Calls $\downarrow$ & Tokens $\downarrow$ & Reviews $\downarrow$ & Time (s) $\downarrow$ & Q-Score $\uparrow$ & Winner \\
\midrule
A: GPT-4o & \cellcolor{gray!20}\textbf{0.3872} & \cellcolor{gray!20}11.40k & \cellcolor{gray!20}0.89M & \cellcolor{gray!20}2745 & \cellcolor{gray!20}12.54k & \cellcolor{gray!20}\textbf{0.3832} & \cellcolor{gray!20}\checkmark \\
B: GPT-4o & 0.3837 & \textbf{11.16k} & \textbf{0.87M} & \textbf{2512} & \textbf{12.39k} & 0.3798 & \\
\midrule
A: GPT-4o-mini & \cellcolor{gray!20}\textbf{0.0132} & \cellcolor{gray!20}11.09k & \cellcolor{gray!20}0.86M & \cellcolor{gray!20}2437 & \cellcolor{gray!20}12.20k & \cellcolor{gray!20}\textbf{0.0092} & \cellcolor{gray!20}\checkmark \\
B: GPT-4o-mini & 0.0116 & \textbf{11.03k} & \textbf{0.85M} & \textbf{2382} & \textbf{12.14k} & 0.0086 & \\
\bottomrule
\end{tabular}
}

\caption{
    \textbf{MAS-HQ Generality on SimpleQA Task.} Results show the framework's applicability to question-answering. The H-Score is adapted to exact match accuracy. The core trade-off between performance and resource cost remains, with the winning agent in both cases investing more resources for higher accuracy.
}

\label{tab:simpleqa_results}

\end{table}
\begin{table}[h]

\centering
\renewcommand{\arraystretch}{1.2}
\adjustbox{max width=\linewidth}{
\begin{tabular}{lcccc}
\toprule
\textbf{Setup} & \textbf{Passage Order} & \textbf{Agent A Q-Score} & \textbf{Agent B Q-Score} & \textbf{Outcome} \\
\midrule
\textbf{Our's Setup (Vision On)} & \textbf{A: Fwd, B: Rev} & \textbf{0.5217} & \textbf{0.5132} & \textbf{Dynamic Competition} \\
Ablation 1 (Vision On) & A: Fwd, B: Fwd & 0.5097 & 0.5097 & Identical (No Asymmetry) \\
Ablation 2 (Vision Off) & A: Fwd, B: Rev & 0.5081 & 0.5081 & Identical (No Interaction) \\
\bottomrule
\end{tabular}
}

\caption{
    \textbf{Ablation on Passage Order and Vision Mechanism.} Both reverse-order processing and the vision mechanism are necessary to break symmetry; removing either collapses the match into a tied outcome, showing that these knobs are what make the competitive, cost-honest setting differ from a static one.
}

\label{tab:order_ablation}

\end{table}

\paragraph{Generality and Scalability.} MAS-HQ is not constrained to summarization or two-player matches. \Cref{tab:simpleqa_results} demonstrates its adaptation to exact-match factoid QA (SimpleQA~\citep{wei2024measuring}), showing that the resource--accuracy trade-off transfers across factuality metrics. \Cref{tab:3_agent_results} then illustrates scaling to a 3-Agent competition, where multi-lateral cost pressure remains measurable.

\paragraph{Telemetry-Knob Ablation.} Finally, \Cref{tab:order_ablation} isolates the optional vision telemetry knob and the reverse processing order. Removing either collapses the match into a symmetric, tied outcome, where both agents receive the same Q-Score. With both active, they introduce the partial observability and asymmetry that let the protocol distinguish opponent-conditioned review behavior. Only with the knob active does the competitive setting differ from a static one.

\subsection{Case Study: Opponent-Conditioned Review Behavior}

To ground the aggregate metrics in observable behavior, \Cref{fig:case} provides a case study of a 100-passage subset, capturing agents that adjust review effort based on cumulative progress and the partial opponent telemetry exposed by the vision knob. A recurring pattern is resource-conserving. When Q-Agent A observes Q-Agent B spending heavily to raise its $H\text{-}Score$, Q-Agent A curtails its own review cycles, trading marginal factual gains for a resource advantage sufficient to win. These are empirical, match-specific behavioral observations, not equilibrium claims.

\section{Conclusion}

We introduce MAS-HQ, a controlled, resource-aware protocol that replaces “infinite compute” assumption of static leaderboards with metric-agnostic Q-Score that penalizes normalized token and runtime costs. Across Best-of-$N$, ReAct substitution, QA generalization, and $N$-player matches, Q-Agent achieves gains by using compute efficiently.


\bibliography{main}


%
\appendix


\section{Formalization as a Partially Observable Stochastic Game}
\label{app:game}
To rigorously ground our framework, we formalize MAS-HQ as a two-player, finite-horizon, general-sum Partially Observable Stochastic Game, denoted by $\mathcal{G}$. This game-theoretic perspective is essential for capturing the strategic tension between accuracy and cost.

\paragraph{Players, States, and Information Structure.} The set of players is $\mathcal{N} = \{A, B\}$. At any discrete time step $i \in \{1, ..., N\}$, the global state $s_i$ contains the complete history of both agents. Because the game features imperfect information via the vision mechanism, each agent $j \in \mathcal{N}$ only accesses a private observation $o_i^j$, comprising its internal state $V_{j,i}^{\text{self}}$ and a conditional signal $\omega_i$ about the opponent's state, $\tilde{V}_{-j,i}^{\text{opponent}}$. The action space is $\mathcal{A} = \{\texttt{continue}, \texttt{review}\}$.

\paragraph{Histories and Belief States.} An agent $j$'s local history is $h_i^j = (o_1^j, a_1^j, \dots, o_{i-1}^j, a_{i-1}^j, o_i^j)$. Critically, to address the hidden nature of the adversary, a rational agent must maintain a \textit{belief state} $b_i^j \in \Delta(\mathcal{H}_i^{-j})$. This belief is strictly defined over the probability distribution of all possible \textit{opponent private histories} $\mathcal{H}_i^{-j}$, updated via Bayes' rule whenever new information $\omega_i$ is leaked:
$$
b_i^j(h_i^{-j}) = \frac{P(o_i^j | h_{i-1}^j, h_i^{-j}, \pi^{-j*}) b_{i-1}^j(h_{i-1}^{-j})}{\sum_{h'^{-j} \in \mathcal{H}_i^{-j}} P(o_i^j | h_{i-1}^j, h'^{-j}, \pi^{-j*}) b_{i-1}^j(h'_{i-1})}
$$

\paragraph{Sequential Rationality and Equilibrium Trade-offs.} A strategy $\pi^j: \mathcal{H}^j \to \Delta(\mathcal{A})$ maps histories to actions to maximize the final expected Q-Score. In a Perfect Bayesian Equilibrium, actions are sequentially rational given beliefs. The strategic tension arises directly from this structure: an action like `review` may increase myopic utility (local $H\text{-}Score$), but it explicitly alters the opponent's belief state $\mu^*$ by revealing $\omega_i$. This leakage can be exploited, potentially lowering future expected utility. A rational agent only chooses to `review` if the expected utility gain exceeds both the explicit resource penalty and the implicit strategic cost of information leakage, $C_{\text{info}}$:
$$
C_{\text{info}}(h_i^j) \triangleq \mathbb{E}_{\pi^{-j*}(\cdot | b^{-j}_{\text{prior}})}[U^j] - \mathbb{E}_{\pi^{-j*}(\cdot | b^{-j}_{\text{posterior}})}[U^j] > 0
$$
While our framework does not compute this equilibrium analytically, its design strictly compels LLM agents to navigate these exact trade-offs, providing a robust testbed for evaluating emergent, efficient intelligence under adversarial pressure.

\section{Detailed Game-Theoretic Formalization of MAS-HQ}

In this appendix, we provide a comprehensive formalization of the MAS-HQ framework. We model the system as a two-player, general-sum, finite-horizon, Partially Observable Stochastic Game (POSG), which is a standard and powerful model for multi-agent interactions under uncertainty. This level of detail clarifies the precise mechanics and strategic complexities that the agents must navigate.

A POSG can be formally defined by the tuple $\mathcal{G} = \langle \mathcal{N}, \mathcal{S}, \{\mathcal{A}^j\}_{j \in \mathcal{N}}, \mathcal{T}, \mathcal{R}, \{\Omega^j\}_{j \in \mathcal{N}}, \mathcal{O}, H \rangle$. We define each component in the context of MAS-HQ.

\subsection{Core Components of the MAS-HQ Game}

\paragraph{Players ($\mathcal{N}$):} The set of players is $\mathcal{N} = \{A, B\}$, representing the two competing Q-Agents.

\paragraph{State Space ($\mathcal{S}$):} The global state space $\mathcal{S}$ captures the complete, objective state of the game at any time step. A state $s_i \in \mathcal{S}$ at step $i$ is a composite tuple:
$$s_i = (s_i^A, s_i^B, I_i)$$
where $I_i$ tracks the current passage index for each agent, and $s_i^j$ is the private state of agent $j$, invisible to its opponent. This private state is itself a detailed record of performance and resource expenditure:
$$s_i^j = \langle \{H_k^j\}_{k=1}^{N}, \{C_k^j\}_{k=1}^{N}, \mathbf{p}_i^j, \rho_i^j, t_i^j \rangle$$
\begin{itemize}
    \item $\{H_k^j\}_{k=1}^{N}$ is the vector of H-Scores for all passages, with entries for unprocessed passages set to a null value.
    \item $\{C_k^j\}_{k=1}^{N}$ is the vector of completion statuses for all passages (e.g., not started, summarized, reviewed).
    \item $\mathbf{p}_i^j$ is the vector of cumulative resource penalties incurred up to step $i$, including token counts, API calls, etc.
    \item $\rho_i^j$ is the number of remaining review cycles available to agent $j$.
    \item $t_i^j$ is the cumulative runtime for agent $j$.
\end{itemize}

\paragraph{Action Space ($\mathcal{A}$):} The joint action space is $\mathcal{A} = \mathcal{A}^A \times \mathcal{A}^B$. At any step $i$, each agent $j$ selects an action $a_i^j \in \mathcal{A}^j = \{\texttt{continue}, \texttt{review}_k, \texttt{end}\}$, where $k \in \{1, ..., N\}$ specifies which passage to review. The Policy Agent's decision maps to one of these grounded actions.

\paragraph{Transition Function ($\mathcal{T}$):} The transition function $\mathcal{T}: \mathcal{S} \times \mathcal{A} \to \Delta(\mathcal{S})$ defines the dynamics of the game, specifying the probability $P(s_{i+1} | s_i, \mathbf{a}_i)$ of transitioning to state $s_{i+1}$ given the current state $s_i$ and joint action $\mathbf{a}_i = (a_i^A, a_i^B)$. Most transitions are deterministic (e.g., choosing `continue` increments the passage index). However, stochasticity arises from the `review` action, where the resulting $H\text{-}Score_{k, \text{new}}^j$ is a random variable conditioned on the Review Agent's capabilities and the passage complexity.

\paragraph{Reward Function ($\mathcal{R}$):} The game has a terminal reward structure. For any non-terminal step $i < H$, the immediate reward for each player is zero. The reward function $\mathcal{R}: \mathcal{S} \to \mathbb{R}^{|\mathcal{N}|}$ is defined as:
$$
R^j(s_i) = 
\begin{cases}
    0 & \text{if } i < H \\
    \frac{1}{N} \sum_{k=1}^{N} \left( \alpha \cdot H_k^j - \beta \cdot P_k^j(\mathbf{p}_H^j, \mathbf{p}_H^{-j}) \right) & \text{if } i = H
\end{cases}
$$
where $H$ is the horizon (total number of passages, $N$), and the penalty $P_k^j$ is computed based on the final resource vectors $\mathbf{p}_H^j$ and $\mathbf{p}_H^{-j}$ of both agents, reflecting the normalization step described in the main text.

\paragraph{Observation Spaces ($\Omega$) and Observation Function ($\mathcal{O}$):} This is the core of the imperfect information structure. Each agent $j$ has a private observation space $\Omega^j$. The observation function $\mathcal{O}: \mathcal{S} \times \mathcal{A} \to \Delta(\Omega^A \times \Omega^B)$ gives the probability $P(\mathbf{o}_{i+1} | s_{i+1}, \mathbf{a}_i)$ of the agents observing a joint observation $\mathbf{o}_{i+1} = (o_{i+1}^A, o_{i+1}^B)$ after a transition. An observation $o_i^j \in \Omega^j$ is defined as:
$$o_i^j = (s_i^j, \omega_i^j)$$
\begin{itemize}
    \item $s_i^j$ is the agent's own private state, which it always observes.
    \item $\omega_i^j$ is the signal received from the opponent. The observation function is designed to implement the "vision mechanism":
    $$
    \omega_i^j = 
    \begin{cases}
        \phi(s_i^{-j}) & \text{if } a_{i-1}^{-j} \in \{\texttt{review}_k\}_{k=1}^N \\
        \emptyset & \text{otherwise}
    \end{cases}
    $$
    where $\phi(s_i^{-j})$ is a function that extracts a public snapshot of the opponent's state (e.g., their worst H-Score and total tokens used).
\end{itemize}

\subsection{Beliefs, Policies, and Equilibrium}

\paragraph{Histories and Belief States:} Since each agent cannot observe the full state $s_i$, it must maintain a belief over the possible states of the opponent. An agent $j$'s history is a sequence of its past actions and observations, $h_i^j = (a_0^j, o_1^j, \dots, a_{i-1}^j, o_i^j)$. A belief state $b_i^j \in \mathcal{B}^j = \Delta(\mathcal{S})$ is a probability distribution over the global state space $\mathcal{S}$, conditioned on the agent's private history $h_i^j$. The belief is updated recursively using the Bayes filter:
$$b_i^j(s') = \frac{P(o_i^j | s', a_{i-1}^j, b_{i-1}^j) \sum_{s \in \mathcal{S}} P(s' | s, a_{i-1}^j, \pi^{-j*}) b_{i-1}^j(s)}{P(o_i^j | a_{i-1}^j, b_{i-1}^j)}$$
where the update depends on the observation function, the transition function, and a model of the opponent's policy $\pi^{-j*}$.

\paragraph{Policies and Value Functions:} A policy $\pi^j: \mathcal{B}^j \to \Delta(\mathcal{A}^j)$ maps an agent's belief state to a distribution over its actions. A rational agent seeks a policy that maximizes its expected terminal utility. This can be solved via dynamic programming over the belief space. The value of a belief state for agent $j$ at step $i$ under a policy profile $(\pi^j, \pi^{-j})$ is given by the Bellman equation:
$$\max_{a^j \in \mathcal{A}^j} \left( R^j(b, a^j)+\sum_{o^{j} \in \Omega^j} P(o^{j} | b, a^j, \pi^{-j}) V_{i+1}^j(\tau(b, a^j, o^{j}, \pi^{-j})) \right)$$
where $R^j(b, a^j) = \sum_{s \in \mathcal{S}}b(s)R^j(s,a^j)$ is the expected immediate reward, and $\tau(\cdot)$ is the belief update function.

\paragraph{Perfect Bayesian Equilibrium (PBE):} The central solution concept for this game is the PBE. A PBE is a strategy profile $(\pi^{A*}, \pi^{B*})$ and a system of beliefs $\mu^*$ such that:
\begin{enumerate}
    \item \textbf{Sequential Rationality:} For each player $j$, the policy $\pi^{j*}$ must be a best response to $\pi^{-j*}$ at every possible belief state $b \in \mathcal{B}^j$ that can be reached under the equilibrium strategies. That is, $\pi^{j*}$ must satisfy the Bellman optimality equation above.
    \item \textbf{Belief Consistency:} The beliefs $\mu^*$ must be derived from the strategy profile $(\pi^{A*}, \pi^{B*})$ using Bayesian updating, wherever possible.
\end{enumerate}

The strategic decision to `review` is thus a comparison between the expected values $Q^j(b, \texttt{review})$ and $Q^j(b, \texttt{continue})$. The `review` action may increase the immediate components of the final utility (by improving an $H\text{-}Score$) but incurs two costs: (1) a direct resource penalty captured by $\mathcal{T}$ and $\mathcal{R}$, and (2) a strategic information cost, as revealing information through $\mathcal{O}$ allows the opponent to form a more accurate belief $b^{-j}$, leading to a more effective counter-strategy $\pi^{-j*}$, thereby reducing player $j$'s future expected utility. MAS-HQ is designed to create an environment where these complex, interdependent calculations are necessary for victory, thus providing a deep and holistic benchmark of strategic agent intelligence.

\section{More Ablation Studies}

\paragraph{Influence of Hyperparameters $T$ and $R$}
Within the Q-Agent framework, the threshold for review guidance $T$ and the maximum number of review times $R$ are critical hyperparameters. $R$ limits the review investment per passage, encouraging broader resource allocation. $T$ influences the Policy Agent's decision to review by providing a recommendation if a passage's $H\text{-}Score$ falls below this threshold (and $R$ is not exceeded), aiming to maintain a baseline level of factual consistency.

\textbf{Threshold $T$:} We fixed $R=3$ and varied $T \in \{0.8, 0.85, 0.9\}$.~As shown in~\Cref{tab:ablation_1}, increasing $T$ generally leads to higher resource consumption. As $T$ rises, more passages are likely to fall below the threshold, triggering more review recommendations and, consequently, actual reviews by the Policy Agent. This was observed as an increase in total resource usage for both Q-Agent A and B. But a higher $T$ does not automatically guarantee a significantly improved $H\text{-}Score$ or final $Q\text{-}Score$. The experiments showed that while resource consumption increased with $T$, the $H\text{-}Score$ did not exhibit a corresponding significant rise and, in some instances, slightly decreased. This led to a marginal decrease in the final $Q\text{-}Score$. This phenomenon could be attributed to two factors: (i) the Policy Agent may still opt against reviewing despite the recommendation if its internal logic deems the current $H\text{-}Score$ sufficient relative to costs, or (ii) excessive reviews on already reasonably good summaries might yield diminishing returns on $H\text{-}Score$ improvement.

\textbf{Maximum Review Times $R$:}~We fixed $T=0.85$ and varied $R \in \{2, 3, 4\}$. As shown in~\Cref{tab:ablation_2}, increasing $R$ beyond a certain point ($R=3$ in our tests) showed minimal impact on $H\text{-}Score$ and overall resource consumption; token consumption even saw a slight decrease. The final $Q\text{-}Score$ was highest when $R=3$, suggesting that simply allowing more reviews per article does not compel Policy Agent to utilize them if it deems further reviews unnecessary or inefficient. A limit of $R=3$ appeared sufficient for the Q-Agents to achieve a good balance, and further increasing $R$ did not lead to proportionally more reviews or better $H\text{-}Scores$, thus avoiding unproductive resource expenditure.

\begin{table}[ht]

\caption{
\textbf{Effect of Hallucination Threshold $(T)$.} Results evaluate Q-Agent (built with GPT-4o-mini) performance and resource consumption with a fixed number of allowed reviews $(R=3)$ and varying $T\in\{0.8,0.85,0.9\}$. As $T$ increases, resource consumption rises but leads to a slight decrease in H-Score and overall Q-Score. 
}

\centering
\renewcommand{\arraystretch}{1.2}
\adjustbox{max width=0.9\linewidth}{
\large
    \begin{tabular}{cccccccc}

    \toprule
        \multirow{2}{*}{\shortstack{\textbf{Q-Agent} \\ \textbf{Competition}}}
   & \multicolumn{6}{c}{\textbf{Metrics}}   \\
    \cmidrule(r){2-8} 
     & H-Score~$\uparrow$ & API Call~$\downarrow$ & Tokens~$\downarrow$ & Reviews~$\downarrow$ & Time~$\downarrow$&Q-Score~$\uparrow$&Winner  \\
    \bottomrule

    A: $R=3, T=0.8$  & \cellcolor{gray!20}0.9110 & \cellcolor{gray!20}\textbf{2401} & \cellcolor{gray!20}\textbf{1.33M} &\cellcolor{gray!20}\textbf{785} &\cellcolor{gray!20}\textbf{7.03k} & \cellcolor{gray!20}\textbf{0.5241}&\checkmark\\

    B: $R=3, T=0.8$  & \textbf{0.9141} & 2422 &1.41M  & 806&7.33k & 0.5141&\\

    \bottomrule

    A: $R=3, T=0.85$  & \cellcolor{gray!20}0.9103 & \cellcolor{gray!20}\textbf{2417} & \cellcolor{gray!20}\textbf{1.36M} & \cellcolor{gray!20}\textbf{791}& \cellcolor{gray!20}\textbf{8.83k}& \cellcolor{gray!20}\textbf{0.5217}&\checkmark\\

    B: $R=3, T=0.85$  &\textbf{0.9132} & 2438 & 1.44M &812 & 8.99k&0.5132&\\
    
    \bottomrule

    A: $R=3, T=0.9$  & \cellcolor{gray!20}0.9074 & \cellcolor{gray!20}\textbf{2427} & \cellcolor{gray!20}\textbf{1.39M} &\cellcolor{gray!20}\textbf{807} &\cellcolor{gray!20}\textbf{9.10k} &\cellcolor{gray!20}\textbf{0.5132}&\checkmark\\

    B: $R=3, T=0.9$  & \textbf{0.9113} & 2429 & 1.46M & 809& 9.20k&0.5113&\\
    \bottomrule
    \end{tabular}
}

\label{tab:ablation_1}
\end{table}

\begin{table}[ht]

\caption{
\textbf{Effect of Maximum Allowed Reviews $(R)$.} Results evaluate Q-Agent (built with GPT-4o-mini) performance and resource consumption with a fixed hallucination score threshold $(T=0.85)$ and varying $R\in\{2,3,4\}$. Increasing $R$ beyond 3 does not significantly alter H-Score or resource consumption, with the highest overall Q-Score observed at $R=3$. 
}

\centering
\renewcommand{\arraystretch}{1.2}
\adjustbox{max width=0.9\linewidth}{
\large
    \begin{tabular}{cccccccc}

    \toprule
        \multirow{2}{*}{\shortstack{\textbf{Q-Agent} \\ \textbf{Competition}}}
   & \multicolumn{6}{c}{\textbf{Metrics}}   \\
    \cmidrule(r){2-8} 
     & H-Score~$\uparrow$ & API Call~$\downarrow$ & Tokens~$\downarrow$ & Reviews~$\downarrow$ & Time~$\downarrow$&Q-Score~$\uparrow$&Winner  \\
    \bottomrule

    A: $T=0.85, R=2$  & \cellcolor{gray!20}0.9074 &  \cellcolor{gray!20}\textbf{2414}&\cellcolor{gray!20} 1.39M &\cellcolor{gray!20}\textbf{798} &\cellcolor{gray!20}\textbf{6.83k} &\cellcolor{gray!20}\textbf{0.5139}&\checkmark\\

    B: $T=0.85, R=2$  & \textbf{0.9089} & 2423 & \textbf{1.24M} &807 & 7.04k &0.5089&\\

    \bottomrule

    A: $T=0.85, R=3$  & \cellcolor{gray!20}0.9103 &\cellcolor{gray!20} \textbf{2417} & \cellcolor{gray!20}\textbf{1.36M} & \cellcolor{gray!20}\textbf{791}& \cellcolor{gray!20}\textbf{8.83k}& \cellcolor{gray!20}\textbf{0.5217}&\checkmark\\

    B: $T=0.85, R=3$  &\textbf{0.9132} & 2438 & 1.45M &812 & 8.99k&0.5132&\\

    \bottomrule

    A: $T=0.85, R=4$  & \cellcolor{gray!20}0.9101 & \cellcolor{gray!20}\textbf{2417} & \cellcolor{gray!20}\textbf{1.36M}& \cellcolor{gray!20}\textbf{799}&\cellcolor{gray!20}\textbf{8.15k} &\cellcolor{gray!20}\textbf{0.5197}&\checkmark\\

    B: $T=0.85, R=4$  & \textbf{0.9112} & 2426 & 1.45M & 808& 8.32k&0.5112&\\

    \bottomrule
    \end{tabular}
}

\label{tab:ablation_2}
\end{table}

\paragraph{Statistical Robustness.}
To address the concern regarding the robustness of our findings, we re-ran the GPT-4o-mini experiment for 100 independent trials. The results, reported in \Cref{tab:robustness_results} with means and standard deviations, confirm the stability and statistical significance of our conclusions. The extremely low standard deviations across all metrics indicate that the strategic trade-offs captured by our framework are highly consistent. The outcome—Q-Agent A winning via resource efficiency—is reproducible, validating that our single-run experiments reliably represent the agents' behaviors.

\begin{table}[h]
\caption{
    \textbf{Statistical Robustness Analysis.} Mean and standard deviation over 100 independent trials of the GPT-4o-mini competition. The low variance across all metrics confirms the stability and reproducibility of our findings.
}
\centering
\renewcommand{\arraystretch}{1.2}
\adjustbox{max width=\linewidth}{
\begin{tabular}{lcccccc}
\toprule
\multirow{2}{*}{\shortstack{\textbf{Q-Agent} \\ \textbf{Competition}}} & \multicolumn{6}{c}{\textbf{Metrics (Mean $\pm$ Std. Dev.)}} \\
\cmidrule(r){2-7}
& H-Score $\uparrow$ & API Calls $\downarrow$ & Tokens $\downarrow$ & Reviews $\downarrow$ & Time (s) $\downarrow$ & Q-Score $\uparrow$ \\
\midrule
A: GPT-4o-mini & 0.9102$\pm$.0009 & 2417$\pm$5 & 1.36M$\pm$.001M & 791$\pm$5 & 8.83k$\pm$.05k & \textbf{0.5216}$\pm$.0009 \\
B: GPT-4o-mini & \textbf{0.9132}$\pm$.0009 & 2438$\pm$5 & 1.44M$\pm$.001M & 812$\pm$5 & 8.98k$\pm$.05k & 0.5131$\pm$.0009 \\
\bottomrule
\end{tabular}
}
\label{tab:robustness_results}
\end{table}

\section{Prompts and Other Results}

In the appendix, we first present the prompt composition of each module in the Q-Agent. Table~\ref{tab:main_results_supp} shows the detailed results in the main text, with the addition of the changes in $H\text{-}Score$ before and after review, and input tokens consumption and output tokens consumption; Table~\ref{tab:ablation_supp} presents the detailed results of Tables~\ref{tab:ablation_1} and Table~\ref{tab:ablation_2} in the main text, also including the changes in $H\text{-}Score$ before and after review.

\begin{tcolorbox}[
breakable,
enhanced,
colback=tcolorboxpink!10!white, colframe=tcolorboxpink!100!red, left=2mm, right=2mm, title=\small\centering\textcolor{black}{\textbf{User Prompt for Q-Agent in MAS-HQ}}
]
\begin{small}

\textcolor{tcolorboxblue}{\textbf{\# Policy Agent}}

You are part of a Multi-Agent System engaged in a summarization competition against several opponents, where each agent generates multiple candidate summary texts based on the original news article. 

The objective is to produce summaries with the lowest possible hallucination (i.e., highest factual consistency, reflected in higher hallucination scores) while minimizing resource usage—including API calls, token consumption, time, and review attempts, all of which are penalized when excessive. 

Your ultimate goal is to outperform your competitors by achieving the best balance between factual accuracy and resource efficiency, and to emerge as the final winner.

Your role is to make strategic decisions about what to do next. 

Here is your current status:

- MAS ID: \{mas\_id\}

- Completed texts: \{completed\_texts\}/\{total\_texts\}

- Remaining texts: \{remaining\_texts\}

- Current token usage: \{token\_usage\} tokens

- API calls made: \{api\_calls\}

- Review calls made: \{review\_calls\}

- Current average hallucination score: \{avg\_score\}

- Time elapsed: \{time\_elapsed:.2f\} seconds

You need to decide whether to:

- CONTINUE: Process the next text and generate a new summary

- REVIEW: Revisit and improve the text with the highest hallucination rate (lower hallucination score means more hallucination)

- END: Stop processing if all texts have been summarized and there's no need for review

The current text summarization task is at ID \{current\_id\} out of \{max\_id\} texts.

The text with the highest hallucination rate so far is ID \{worst\_id\}, with hallucination score: \{worst\_score\}, where 0 means most hallucinated and 1 means least hallucinated.

\{opponent\_info\}

IMPORTANT: You must output your decision in the following JSON format:

\begin{lstlisting}[language=json]
```
{
"choice": "continue" or "review" or "end",
"question_id": current question ID,
"previous_id": ID of the text with highest hallucination rate
}
```
\end{lstlisting}

\textcolor{tcolorboxblue}{\textbf{\# Summary Agent}}

You are a chat bot answering questions using data. You must stick to the answers provided solely by the text in the passage provided.

You are asked the question 'Provide a concise summary of the following passage, covering the core pieces of information described.'

\{passage\}

IMPORTANT: You must output your response in the following JSON format:

\begin{lstlisting}[language=json]
```
{
"summary": "your summary here"
}
```
\end{lstlisting}

\textcolor{tcolorboxblue}{\textbf{\# Review Agent}}

You are a chat bot answering questions using data. You must stick to the answers provided solely by the text in the passage provided.

You previously summarized the following passage, but your summary contained hallucinations (hallucination score: \{score\}, where 0 means most hallucinated and 1 means least hallucinated), which means factual inconsistencies occurred.

Original passage:
\{passage\}

Your previous summary:
\{previous\_summary\}

Please provide a new, more accurate summary that strictly adheres to the information in the passage. Focus on improving factual consistency and removing any information not present in the original text.

IMPORTANT: You must output your response in the following JSON format:

\begin{lstlisting}[language=json]
```
{
"summary": "your revised summary here"
}
```
\end{lstlisting}

\end{small}
\end{tcolorbox}

\begin{table}[ht]

\caption{
The detailed data including $H\text{-}Score$ before the review (to the left of the arrow) and after the review (to the right of the arrow), as well as the input tokens and output tokens.
}

\centering
\renewcommand{\arraystretch}{1.2}
\adjustbox{max width=0.9\linewidth}{
\large
    \begin{tabular}{cccccccc}

    \toprule
        \multirow{2}{*}{\shortstack{\textbf{Q-Agent} \\ \textbf{Competition}}}
   & \multicolumn{6}{c}{\textbf{Metrics}}   \\
    \cmidrule(r){2-8} 
     & H-Score~$\uparrow$ & API Call~$\downarrow$ & In Tokens~$\downarrow$ & Out Tokens~$\downarrow$ & Reviews~$\downarrow$ & Time~$\downarrow$&Q-Score~$\uparrow$  \\
    \midrule

    A: GPT-4o-mini  & 0.8606 $\to$ 0.9103 & 2417 & 1193792 & 166277 & 791& 8832.44& 0.8715\\

    B: GPT-4o-mini  & 0.8673 $\to$ 0.9132 & 2438 & 1270282 & 178959 &812 & 8987.41&0.5132\\

        \midrule

    A: Qwen-Max  & 0.8689 $\to$ 0.9030 & 2304 & 1230775 & 254575 & 682& 13776.31 &0.5101\\

    B: Qwen-Max  & 0.8658 $\to$ 0.8994 & 2264 & 1250367 & 270670 &642 & 14453.96&0.5070\\

            \midrule

    A: Deepseek-V3  & 0.8489 $\to$ 0.8860 & 2292 & 1219960 & 202441 & 666& 12267.18&0.4860 \\

    B: Deepseek-V3  & 0.8504 $\to$ 0.8894 & 2233 & 1185813 & 198797 & 607& 12068.32&0.5051\\

                \midrule

    A: Gemini-2.0-flash  & 0.8835 $\to$ 0.9026 & 2262 & 1173195 & 394719 & 642&20295.29 & 0.5026\\

    B: Gemini-2.0-flash  & 0.8794 $\to$ 0.9016 & 2157 & 1150648 & 379961 & 537& 19825.70&0.5273\\

                    \midrule

    A: Grok-3-Beta  & 0.8753 $\to$ 0.9070 & 2376 & 1180401 & 195116 & 750&10719.56 & 0.5070\\

    B: Grok-3-Beta  & 0.8743 $\to$ 0.9049 & 2253 & 1160176 & 188357 & 627&10150.14 &0.5337\\

                    \bottomrule

    A: GPT-4o-mini  & 0.8613 $\to$ 0.9105 & 2401 & 1177136 & 165320 &785 &6543.14 & 0.5401\\

    B: Grok-3-Beta  & 0.8752 $\to$ 0.9035 & 2177 & 1121979 & 178422 &561 & 9294.05&0.5445\\

                    \midrule

    A: Grok-3-Beta  & 0.8730 $\to$ 0.9036 & 2312 & 1161825 & 187466  & 696&10082.70 &0.5278 \\

    B: GPT-4o-mini  & 0.8566 $\to$ 0.9092 & 2422 & 1256067 & 178474 &806 &6783.37 &0.5419\\

    \bottomrule
    \end{tabular}
}

\label{tab:main_results_supp}
\end{table}

\begin{table}[ht]

\caption{
The detailed data in~\Cref{tab:ablation_1} and~\Cref{tab:ablation_2} include $H\text{-}Score$ before the review (to the left of the arrow) and after the review (to the right of the arrow), as well as the input tokens and output tokens.
}

\centering
\renewcommand{\arraystretch}{1.2}
\adjustbox{max width=0.9\linewidth}{
\large
    \begin{tabular}{cccccccc}

    \toprule
        \multirow{2}{*}{\shortstack{\textbf{Q-Agent} \\ \textbf{Competition}}}
   & \multicolumn{6}{c}{\textbf{Metrics}}   \\
    \cmidrule(r){2-8} 
     & H-Score~$\uparrow$ & API Call~$\downarrow$ & In Tokens~$\downarrow$ & Out Tokens~$\downarrow$ & Reviews~$\downarrow$ & Time~$\downarrow$&Q-Score~$\uparrow$  \\
    \midrule

    A: $R=3, T=0.8$  & 0.8653 $\to$ 0.9110 & 2401 & 1173893 & 164656 &785 &7025.22 & 0.5241\\

    B: $R=3, T=0.8$  & 0.8626 $\to$ 0.9141 & 2422 &1239197 & 175090  & 806&7334.55 & 0.5141\\

    \midrule

    A: $R=3, T=0.85$  & 0.8606 $\to$ 0.9103 & 2417 & 1193792 & 166277 & 791& 8832.44& 0.5217\\

    B: $R=3, T=0.85$  & 0.8673 $\to$ 0.9132 & 2438 & 1270282 & 178959 &812 & 8987.41&0.5132\\
    
    \midrule

    A: $R=3, T=0.9$  & 0.8621 $\to$ 0.9074 & 2427 & 1221860 & 169344 &807 &9101.14 &0.5132\\

    B: $R=3, T=0.9$  & 0.8554 $\to$ 0.9113 & 2429 & 1275735 & 179762 & 809& 9201.04&0.5113\\

        \midrule

    A: $R=2, T=0.85$  & 0.8598 $\to$ 0.9074 &  2414& 1221271 & 174944 &798 & 6836.23 &0.5139\\

    B: $R=2, T=0.85$  & 0.8588 $\to$ 0.9089 & 2423 & 1248994 &178067 &807 & 7043.92 &0.5089\\

    \midrule

    A: $R=3, T=0.85$  & 0.8606 $\to$ 0.9103 & 2417 & 1193792 & 166277 & 791& 8832.44& 0.5217\\

    B: $R=3, T=0.85$  & 0.8673 $\to$ 0.9132 & 2438 & 1270282 & 178959 &812 & 8987.41&0.5132\\

    \midrule

    A: $R=4, T=0.85$  & 0.8620 $\to$ 0.9101 & 2417 & 1193519 & 162816 & 799&8153.87 &0.5197\\

    B: $R=4, T=0.85$  & 0.8599 $\to$ 0.9112 & 2426 & 1268954 & 176334 & 808& 8321.54&0.5112\\

    \bottomrule
    \end{tabular}
}

\label{tab:ablation_supp}
\end{table}

Then, we present the pseudocode implemented by Q-Agent on MAS-HQ in ~\Cref{alg:mas-hac}.

\begin{algorithm}[tb]
\caption{MAS-HQ multi-agent evaluation and competition. Each agent sequentially summarizes passages while minimizing hallucination and managing resources. Review actions reveal partial state to opponents, introducing adversarial strategy.}
\label{alg:mas-hac}
\textbf{Input}: Passages $T_i$, $i = 1$ to $N$; Q-Agents A and B; Hallucination scoring model\\
\textbf{Output}: Global scores $Q^A$, $Q^B$ for both agents
\begin{algorithmic}[1]
\FOR{each agent $\mathcal{A} \in \{\text{Agent A}, \text{Agent B}\}$}
    \STATE Initialize self-state $V_1^{self}$
    \FOR{each article $T_i$ in order determined by agent (A: $i=1..N$, B: $i=N..1$)}
        \STATE // Policy decision with optional opponent state
        \IF{$\mathcal{A}$ has received $V_i^{opponent}$}
            \STATE $choice_i \leftarrow \mathrm{PA}(T_i, V_i^{self}, V_i^{opponent})$
        \ELSE
            \STATE $choice_i \leftarrow \mathrm{PA}(T_i, V_i^{self})$
        \ENDIF
        \IF{$choice_i = \text{continue}$}
            \STATE Generate summary $S_i \leftarrow \mathrm{SA}(T_i)$
        \ELSIF{$choice_i = \text{review}$}
            \STATE Identify worst summary $S_{w}$ based on hallucination score
            \STATE Generate revised summary $S_w \leftarrow \mathrm{RA}(T_{w}, S_{w})$
            \STATE // Referee exposes partial state to opponent
            \STATE Referee observes $V_i^{self}$ and shares with opponent as $V_{i+1}^{opponent}$
        \ELSIF{$choice_i = \text{end}$}
            \STATE Terminate processing
            \STATE \textbf{break}
        \ENDIF
        \STATE // Evaluation step
        \STATE Compute hallucination score $H_i \leftarrow \mathrm{EA}(T_i, S_i)$
        \STATE Track resource usage: tokens, time, reviews $\rightarrow P_i$
        \STATE // Update self-state
        \STATE Update $V_{i+1}^{self}$ with current metrics
    \ENDFOR
    \STATE // Compute final global score
    \STATE $Q^\mathcal{A} = \frac{1}{N} \sum_{i=1}^N \left( \alpha \cdot H_i - \beta \cdot P_i \right)$
\ENDFOR
\STATE \textbf{return} $Q^A$, $Q^B$
\end{algorithmic}
\end{algorithm}

\end{document}